\documentclass[sigconf]{acmart}
\usepackage{makecell}
\usepackage{multirow}

\usepackage{amssymb}
\usepackage{pifont}% http://ctan.org/pkg/pifont
\newcommand{\cmark}{\ding{51}}%
\newcommand{\xmark}{\ding{55}}%
\AtBeginDocument{%
  }

\acmConference[ACM MM '25]{Make sure to enter the correct
  conference title from your rights confirmation email}{October 27--31,
  2025}{Dublin, Ireland}
\setcopyright{acmlicensed}
\copyrightyear{2025}
\acmYear{2025}
\acmDOI{XXXXXXX.XXXXXXX}
\acmConference[Conference acronym 'ACM MM]{Make sure to enter the correct
  conference title from your rights confirmation email}{October 27--31,
  2025}{Dublin, Ireland}
\acmISBN{978-1-4503-XXXX-X/2018/06}

\begin{document}

%%
%% The "title" command has an optional parameter,
%% allowing the author to define a "short title" to be used in page headers.
\title{LEHA-CVQAD: Dataset To Enable Generalized Video Quality Assessment of Compression Artifacts}

%%
%% The "author" command and its associated commands are used to define
%% the authors and their affiliations.
%% Of note is the shared affiliation of the first two authors, and the
%% "authornote" and "authornotemark" commands
%% used to denote shared contribution to the research.

% TODO order of authors
% TODO affiliations
\author{Aleksandr Gushchin}
\authornote{Both authors contributed equally to this research.}
\email{alexander.gushchin@graphics.cs.msu.ru}
\orcid{0002-4055-7394}
\affiliation{%
  \institution{ISP RAS Research Center for Trusted Artificial Intelligence}
  \institution{Lomonosov Moscow State University}
  \city{Moscow}
  \country{Russia}
}
\author{Maksim Smirnov}
\authornotemark[1]
\email{mavlsmirnov@edu.hse.ru}
\affiliation{%
  \institution{HSE}
  \city{Moscow}
  \country{Russia}
}

\author{Dmitriy S. Vatolin}
\email{dmitriy@graphics.cs.msu.ru}
\affiliation{%
  \institution{MSU Institute for Artificial Intelligence}
  \institution{Lomonosov Moscow State University}
  \city{Moscow}
  \country{Russia}
}

\author{Anastasia Antsiferova}
\email{aantsiferova@graphics.cs.msu.ru}
% \email{TODO}
\affiliation{%
  \institution{ISP RAS Research Center for Trusted Artificial Intelligence}
  \institution{MSU Institute for Artificial Intelligence}
  \city{Moscow}
  \country{Russia}}

%%
%% By default, the full list of authors will be used in the page
%% headers. Often, this list is too long, and will overlap
%% other information printed in the page headers. This command allows
%% the author to define a more concise list
%% of authors' names for this purpose.
\renewcommand{\shortauthors}{Antsiferova et al.}

%%
%% The abstract is a short summary of the work to be presented in the
%% article.
\begin{abstract}
We propose the LEHA-CVQAD (Large-scale Enriched Human Annotated) dataset, which comprises 6,240 clips for compression-oriented video quality assessment. 59 source videos are encoded with 186 codec-preset variants, $\approx1.8$M pairwise, and $\approx1.5$k MOS ratings are fused into a single quality scale; part of the videos remains hidden for blind evaluation. We also propose Rate-Distortion Alignment Error (RDAE), a novel evaluation metric that quantifies how well VQA models preserve bitrate-quality ordering, directly supporting codec parameter tuning. Testing IQA/VQA methods reveals that popular VQA metrics exhibit high RDAE and lower correlations, underscoring the dataset’s challenges and utility. The open part and the results of LEHA-CVQAD are available at \url{https://aleksandrgushchin.github.io/lcvqad/}
\end{abstract}

%%
%% The code below is generated by the tool at http://dl.acm.org/ccs.cfm.
%% Please copy and paste the code instead of the example below.
%%


%% 
%% Keywords. The author(s) should pick words that accurately describe
%% the work being presented. Separate the keywords with commas.
%% A "teaser" image appears between the author and affiliation
%% information and the body of the document, and typically spans the
%% page.

% \begin{teaserfigure}
%   \includegraphics[width=\textwidth]{imgs/sampleteaser}
%   \caption{Seattle Mariners at Spring Training, 2010.}
%   \Description{Enjoying the baseball game from the third-base
%   seats. Ichiro Suzuki preparing to bat.}
%   \label{fig:teaser}
% \end{teaserfigure}

\received{30 may 2025}
% \received[revised]{12 March 2009}
% \received[accepted]{5 June 2009}

%%
%% This command processes the author and affiliation and title
%% information and builds the first part of the formatted document.
\maketitle

\section{Introduction}
Video streaming accounts for the most significant portion of global internet traffic. Due to escalating data transmission costs, providers face a need to optimize compression efficiency, seeking a balance between reducing bandwidth and maintaining perceptual quality. This optimization relies fundamentally on video quality assessment (VQA) metrics, which guide codec parameter selection by quantifying tradeoffs between bitrate and visual fidelity. %The importance of quality measurement intensifies with user-generated content (UGC), where ``in-the-wild'' distortions from casual capture devices interact unpredictably with compression artifacts. 

VQA metrics fall into full-reference (FR), which requires pristine source video, and no-reference (NR), which can operate without reference clips. Primarily, FR metrics are used for codec development and guidance in codec tuning. NR metrics are used to assess user-generated content (UGC), which represents ``in-the-wild'' distortions from casual capture devices. 
Current challenges in developing VQA metrics stem from the limitations of publicly accessible datasets. Public compressed video datasets lack the scale and diversity of artifacts, particularly those provided by different codecs of the same compression standard. 
Recent datasets, such as CVQAD~\cite{NEURIPS2022_59ac9f01}, partially mitigate these issues. However, CVQAD does not provide the mean opinion score (MOS) ground truth labels that enable convenient VQA metric training. It also underrepresents authentic distortions of UGC and distortions provided by recent codecs.
Thus, there is a need for large-scale, subjectively labeled VQA datasets that capture the complex, authentic, and synthetic distortions present in real-world videos. Such datasets are essential not only for the development of robust NR and FR VQA metrics but also for advancing the scientific understanding of perceptual video quality. In this paper, we introduce a \textit{Large-scale Enriched Human Annotated CVQAD (LEHA-CVQAD) dataset}, designed to overcome these limitations through three innovations:
\begin{itemize}
    \item {\textbf{Multi-modal quality labels} by three-stage subjective methodology: (i) pairwise ranking within the same content, (ii) intra-content MOS collection, and (iii) a novel method that unifies pairwise and MOS scores onto a single quality scale.}
    \item {\textbf{Content and codec diversity}: LEHA-CVQAD combines professionally produced material with diverse UGC, capturing both authentic artifacts and a broad spectrum of controlled compression distortions. Its compression covers 180+ codecs of different standards, including AV1 and VVC, yielding a total of 6,240 distorted videos.}
    \item {\textbf{Structured evaluation}: LEHA-CVQAD is divided into open and hidden parts to prevent benchmark overfitting, containing 1,963 and 4,277 videos, respectively. We launched 100 IQA/VQA metrics on the proposed dataset and reported the benchmark results on the website~\footnote{\url{https://videoprocessing.ai/benchmarks/video-quality-metrics_both.html}}.}
\end{itemize}
%metric
In addition to the dataset, we propose a new evaluation metric that quantifies how accurately a VQA model preserves quality-bitrate ordering. The Rate-Distortion Alignment Error (RDAE) evaluates the accumulated divergence between a VQA model’s predicted rate-distortion (RD) curve and ground-truth subjective RD curves (Fig.~\ref{fig:example}). \textit{RDAE directly answers a key engineering question: ``How many bits are wasted if we optimize codec parameters using this metric?''}. By penalizing both rank inconsistencies and absolute quality miscalibrations, RDAE provides actionable insights for codec developers, helping to improve bandwidth utilization.
\begin{table*}
  \caption{Summary of subjective video quality datasets and our new dataset.}
  \label{tab:videodatasets}
  \scalebox{0.85}{%
  \begin{tabular}{lccccccccccc}
    \toprule
    Dataset (Year) & \makecell{Source\\type} & \makecell{Orig. / Dist. \\ videos} & \makecell{Compression\\(Codecs)} & CRF & \makecell{Subjective \\ framework} & \makecell{Subjective \\ data}  & \makecell{Ans. per video/\\Subjects} & Access \\
    \midrule

     LIVE-VQA \cite{livevqa} (2008) & Raw  & 10/160 & MPEG-2, AVC (2) & 4 & In-lab & DMOS & 29/38 & \xmark \\
     EPFL-PoliMI \cite{epfl} (2009) & Pristine  & 12/156 & AVC(1) & 1 & In-lab & MOS & 34/40 & \xmark \\
    % VQEG-HDTV (2010) & Pristine  & 49/740 & MPEG-2, AVC(-/-) & In-lab & RAW & 24/120 & \cmark \\
    % IVP (2011) & Raw  & 10/138 & MPEG-2, AVC, DIRAC (3/3) & In-lab & DMOS & 35/42 & \xmark \\
     TUM 1080p50 \cite{keimel2012tum}  (2012) & Raw  & 25/1,560 & AVC(1) & 4 & In-lab & MOS & 21/21 & \xmark \\
    % LIVE Mobile (2012) & Pristine  & 10/200 & AVC(1/4) & In-lab & DMOS & 47/47 & \cmark \\
     CSIQ \cite{csiq}  (2014) & Raw  & 12/228 & \makecell{MJPEG, AVC,\\HEVC, SNOW (4)} & 3 & In-lab & DMOS & -/35 & \xmark \\
     MCL-V \cite{lin2015mcl} (2015) & Raw  & 12/108 & AVC(1) & 4 & In-lab & MOS & 32/45 & \xmark \\
     %MCL-JCV (2016) & Pristine  & 30/1560 & AVC(1/51) & In-lab & RAW-JND & 50/150 & \cmark \\
     VideoSet \cite{videoset} (2017) & Raw & 220/45,760 & AVC (1) & 51 & In-lab & RAW-JND & >30/800 & \cmark  \\     
     %LIVE-NFLX-II (2018) & Raw & 15/420 & ? & In-lab & MOS & 23/65 & \cmark \\
     KUGVD \cite{kugvd}  (2019)  & Raw & 6/90 & AVC (1) & 5 & In-lab & MOS & -/17 &  \cmark \\ 
     LIVE-YT-HFR \cite{liveythfr} (2021) & Raw  & 16/480 & VP9 (1) & 5 & In-lab & MOS, DMOS & 223/85 &  \cmark \\
     BVI-CC \cite{bvicc} (2022) & Raw & 9/306 & HEVC, VVC, AV1 (3) & 4 & In-lab & DMOS & 62/62 &  \cmark \\ 
     % GamingHDRVSET (2019) & Pristine (Gaming) & 6/90 & AVC, HEVC, VP9, AV1 (4/4) & In-lab & MOS & -/17 & \cmark \\ 
     %Tencent GVD (2022) & Pristine (Gaming)  & 150/1293 & ? & In-lab & MOS & -/19 & \xmark \\
     CVD-2014 \cite{cvd2014} (2014) & UGC  & 5/234 & – & - & In-lab & MOS & 30/210 &  \cmark\\
     LIVE-Qualc. \cite{ghadiyaram2017capture} (2016) & UGC  & 54/208 & – & - & In-lab & MOS & 39/39 & \cmark\\
     KoNViD-1k \cite{konvid} (2017) & UGC  & 0/1,200 & – & - & Crowd & MOS & 114/642 &  \cmark\\
     LIVE-VQC \cite{livevqc} (2018) & UGC  & 0/585 & – & -& Crowd & MOS & 240/4,776 &  \cmark\\ 
     YouTube-UGC \cite{ytugc} (2019) & UGC & 0/1,380 & – & - & Crowd & MOS & 123/>8,000 &  \cmark\\
    % FlickrVid-150k (2019) & UGC  & 0/153841 & – & Crowd & MOS & - & 5 & \xmark \\
     LSVQ \cite{lsvq} (2020) & UGC & 0/38,811 &  – & - & Crowd & MOS & 35/6,284 & \cmark\\
     %Tele-VQA (2022)  & UGC & 0/2320 & – & In-lab & MOS & 526 & 34 & \xmark\\
     Maxwell \cite{maxwell} (2023)  & UGC & 0/4,543 & – & - & In-lab & MOS & 440/35 &  \cmark\\
  %   LIVE-YT-GVQA ( 2023) & UGC & 0/600 & – & In-lab & MOS & 31/61 & \xmark \\
     
     UGC-VIDEO \cite{ugcvid} (2020) & UGC & 50/550 & AVC, HEVC (2) & 5 & In-lab & DMOS & 30/30 &  \xmark \\
     LIVE-WC \cite{livewc} (2020) & UGC & 55/275 & AVC (1) & 4 & In-lab & MOS & 40/40 &  \xmark \\
     YT-UGC+ \cite{ytugcplus} (2021) & UGC & 189/567 & VP9 (1) & 3 & In-lab & DMOS & 30/- & \cmark \\
     TaoLive \cite{zhang2023md} (2023) & UGC & 418/3,762 & HEVC (1) & 8  & In-lab & MOS & 44/44 & \cmark \\
     KVQ \cite{lu2024kvq} (2024) & UGC & 600/4,200 & Hidden (1) & 6 & In-lab & MOS, Rank & 15/15 &  \cmark \\
     \midrule
     CVQAD \cite{antsiferova2022video} (Open) (2022) & \makecell{Raw\\+UGC} & 36/1,022 &  \makecell{AVC, HEVC, VVC,\\VP9, AV1 (32)} & 3 & Crowd & Rank & 943/10,800 &  \cmark \\
     CVQAD \cite{antsiferova2022video}  (Full) (2022) & \makecell{Raw\\+UGC} & 36/2,486 &  \makecell{AVC, HEVC, VVC,\\VP9, AV1 (83)} & 3 & Crowd & Rank & 909/10,800 &  Bench. \\
     \midrule
     \makecell{\textbf{LEHA-CVQAD}\\(Open)} (2025) & \makecell{Raw\\+UGC} & 59/1,963 &  \makecell{AVC, HEVC, VVC,\\VP9, AV1 (57)} & 3 & Crowd & \makecell{Rank (BT, ELO)\\MOS, DMOS} & 875/>15,000 & \cmark \\
     \makecell{\textbf{LEHA-CVQAD}\\(Full)} (2025) & \makecell{Raw\\+UGC}  & 59/6,240 & \makecell{AVC, HEVC, VVC,\\VP9, AV1, etc. (186)} & 3 & Crowd & \makecell{Rank (BT, ELO)\\MOS, DMOS} & 608/>15,000 &  Bench. \\
     
  \bottomrule
\end{tabular}
}
\end{table*}

\section{Related Work}
% \textbf{Video compression datasets}.
Subjective video quality datasets can be divided into three types: legacy datasets comprising raw, professionally generated videos that are synthetically distorted, authentic UGC datasets, and their combination. Legacy datasets \cite{livevqa, epfl, keimel2012tum, csiq, mclv, videoset, kugvd, liveythfr, bvicc} contain distortions applied systematically to pristine reference videos. %These distortions may include various forms of compression, transmission impairments, video processing artifacts, and degradations, such as Gaussian blur. 

These datasets include various forms of compression but exhibit limited content diversity. The exclusive use of raw (uncompressed) source videos yields high-quality content that lacks the authentic distortions typically found on social media platforms.
In contrast, UGC datasets \cite{konvid, ytugc, lsvq, livevqc, maxwell, ghadiyaram2017capture, cvd2014} naturally exhibit a wide range of authentic distortions arising from diverse devices, codecs, editing tools, and user behaviors. The drawback of UGC datasets is the absence of original raw videos, which makes them inapplicable for training FR VQA metrics. Moreover, UGC videos usually undergo further recompression on streaming platforms. %Currently, UGC datasets are becoming increasingly prevalent in recent research, reflecting the uncontrolled environment of today's video production. Frequently occurring distortion types related to video capture include insufficient color representation, overexposure, underexposure, autofocus-related distortions, unsharpness, and stabilization-related distortions, such as shakiness. 
The hybrid datasets \cite{ugcvideo, livewc, zhang2023md, ytugcplus, zhang2023md, lu2024kvq, antsiferova2022video} combine the qualities of both legacy and UGC datasets, aiming to capture a broader range of distortions. In these datasets, authentic user-generated videos are processed with additional synthetic distortions, creating a more comprehensive set that includes both natural impairments and systematically controlled degradations. %These developments foster the advancement of VQA technologies and the creation of more robust and perceptually relevant quality assessment tools. 
Some of these datasets are not publicly available, which hinders their use for experimental purposes.

The subjective assessment protocol is an essential factor for VQA datasets. Humans have mostly annotated early datasets in a lab environment. This allowed for quality assessment under controlled conditions. However, in-lab assessment is costly and time-consuming. At least 15–18 ratings per video are recommended for statistically stable results, which limits the size of in-lab-labelled datasets \cite{limits}. With the increasing demand for dataset sizes, crowdsourcing has become an affordable and effective method, albeit with downsides, including reduced control over the environment and annotation quality. Therefore, careful quality control must be considered to ensure the production of high-quality results. Most existing datasets provide MOS ratings, which are raw subjective quality scores averaged across subjects for each video. 
% The Bradley-Terry probabilistic model solves this task by computing a win-count matrix, where each entry reflects the number of times one method was preferred over another. Latent quality scores are then estimated by maximizing the likelihood with one method fixed as a reference for identifiability. %To ensure robustness, regularization was applied, and the optimization was performed via a constrained optimizer.
% The Elo rating system is a method for calculating the relative skill levels of players, which has been widely adopted in LLM evaluations.  %The Elo rating system can handle a large number of competitive videos efficiently and provides a clear, unique ranking for all of them. 
%Nevertheless, the crowdsourced subjective procedure is especially worth considering today, as deep learning approaches outperform classical methods in many tasks and are known to be robust to noisy labels. Due to the size of our dataset, we employ a crowdsourcing subjective environment.

Table~\ref{tab:videodatasets} summarizes the existing datasets of annotated compressed videos. The majority of legacy datasets with raw content primarily feature videos encoded with a limited selection of codecs, mostly AVC. 

The most extensive collections of hybrid synthetic and authentic distortions are TaoLive~\cite{zhang2023md}, KVQ~\cite{lu2024kvq}, CVQAD~\cite{NEURIPS2022_59ac9f01}.
The first two do not have a sufficient number of subjective ratings per video to establish statistically reliable ground-truth quality scores. They also support only a limited range of compression standards, thereby failing to represent the vast diversity of encoder artifacts encountered in contemporary industry practices.

\section{Video Dataset Collection}
\textbf{Source videos selection}.
First, we collected a set of candidate videos for further sampling. We downloaded 25,562 FullHD pristine videos from \url{www.vimeo.com} and \url{https://media.xiph.org/}, utilizing a set of simple words as search keywords. To minimize the presence of artifacts, only videos with a minimum bitrate of 20 Mbps were included in the candidate pool. All downloaded videos were subsequently transcoded to the YUV 4:2:0 chroma subsampling format. The videos did not contain any shot transitions or audio components. We also added videos from YouTube UGC to increase the fraction of user-generated content in the candidate dataset. All candidate videos were distributed under the CC BY or CC0 licenses. Second, we sampled a final reference video set from our candidate set. To enable content diversity, we implemented a clustering strategy that considers spatial and temporal complexity following \cite{siti}.%, spatial complexity was estimated by averaging the ratio of x264-encoded I-frame sizes to the corresponding uncompressed frame sizes, while temporal complexity was defined as the ratio of the mean P-frame size to the mean I-frame size. 
Subsequently, we employed the K-means algorithm \cite{lloyd1982least} to partition the dataset into 59 clusters. For each cluster, we randomly selected up to ten candidate videos and then manually reviewed them to choose a single video. The choice was motivated by ensuring diverse and uniform coverage of video genres. The final set includes the following content: Sports, Gaming, Nature, Interviews and Television Clips, Animation, Vlogs, Advertisements, Music Videos, Water Surfaces, Face Close-Ups, and UGC.  %Many genres exhibit distinct visual properties. For example, animated sequences are characterized by precise edges and simplified color palettes. Sports videos typically feature rapid motion against relatively uncomplicated backgrounds, and horror movies often contain numerous dark scenes. Furthermore, close-up facial shots are ubiquitous in genres such as drama and are particularly relevant, given that human faces are recognized as regions of high visual salience, readily capturing viewer attention. 
%The prevalence of these types of content in streaming platforms makes the inclusion of diverse kinds of content essential to the representativeness of the database. 

\textbf{Encoding source videos}.
To cover a diverse spectrum of compression artifacts, the selected videos were encoded using a variety of codecs that implement different compression standards, including but not limited to H.264/AVC and HEVC/H.265, AV1, VP9, and VVC/H.266. For each reference video, we selected constant-rate factor (CRF) by sampling uniformly from rate-distortion curves of SSIM (6Y:1U:1V). Three distinct presets with three target bitrates (1,000 kbps, 2,000 kbps, and 4,000 kbps) were employed to augment artifact diversity further: the fastest preset enabled real-time encoding at approximately 30 frames per second (FPS), the medium preset operated at 5 FPS, and the slowest preset, offering the highest quality, encoded at 1 FPS. It should be noted that not all videos were encoded with every available codec. 
% добавил выше
% Each video was compressed at three target bitrates: 1,000 kbps, 2,000 kbps, and 4,000 kbps—using variable bitrate (VBR) mode where supported; otherwise, quantization parameter (QP) or constant rate factor (CRF) values were selected to yield the desired average bitrates. 

Notably, industry guidance for streaming services suggests a recommended bitrate ceiling of 4,500–8,000 kbps for FullHD content \cite{IBM_recommended_bitrates, Twitch_recommended_bitrates, YouTube_recommended_bitrates}. In this study, higher target bitrates were intentionally excluded, as compression artifacts become largely imperceptible under such conditions, thereby limiting the relevance of subjective quality assessments. We used a total of 186 codecs (including diverse presets) to develop the final dataset. It resulted in 6,240 videos being distorted due to compression.

\textbf{Dataset division}.
%Most recent VQA metrics are evaluated on publicly available datasets, which raises concerns about potential overfitting or tuning to these benchmarks. 
To enable a fair assessment of future metrics' performance on our dataset, we share only a subset of it. The open part of the dataset contains videos compressed by open-source codecs. At the same time, the hidden set includes videos compressed using proprietary codecs, making evaluation of it especially important for real-world applications of quality metrics. %For the open set, we selected 57 codecs and used all 59 reference videos. As a result, the open set contains 1,963 videos out of the full dataset of 6,420 videos.

\section{Subjective Scores Collection}
%As the number of rating options increases in subjective quality assessment tasks, maintaining score consistency and reliability across multiple assessors becomes increasingly challenging. 
To make LEHA-CVQAD helpful in benchmarking and training new metrics, we supply it with two types of subjective scores: MOS and ranking scores from pairwise comparisons. Pairwise comparisons proved to be a straightforward approach to quality assessment. Collected votes can be converted into ranking quality scores using the Bradley-Terry model~\cite{bradleyterry} or Elo system~\cite{elo}. However, pairwise comparison can be conducted only for videos compressed from the same raw source. The resulting relative ranking scores are incomparable across the whole video set, which limits the training of more stable and generalizable VQA models. Therefore, we also computed MOS values as a standardized and content-independent measure of quality. Adding MOS enabled comparability of evaluation results both within our dataset and with existing ones.

%The interpretation of a given rating can vary substantially between different participants, and even for the same individual, decisions may fluctuate over time due to cognitive biases or fatigue. Such variability can undermine the stability and interpretability of absolute quality scores, such as Mean Opinion Score (MOS). To address these challenges, we employed a pairwise comparison approach as the primary method for subjective testing. By restricting assessors to making only relative quality judgments between pairs of stimuli, the pairwise comparison method reduces ambiguity and the cognitive burden on participants, yielding more consistent subjective quality assessments.

\begin{figure*}
  \includegraphics[width=\textwidth]{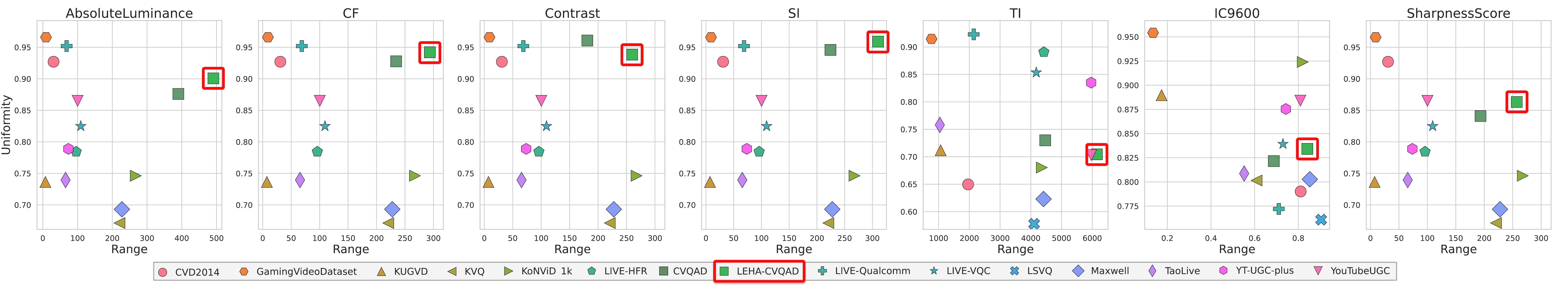}
  \caption{Statistics (Range, Uniformity) of various complexity metrics for LEHA-CVQAD and other datasets.}
  \Description{LEHA-CVQAD complexity comparison with other datasets.}
  \label{fig:complexity}
\end{figure*}

\textbf{Comparison framework}. All subjective data were collected through the crowdsourcing service \url{www.subjectify.us}. Participants completed the experiments in a web browser that automatically switched to full-screen playback; videos were presented at FullHD resolution to avoid scaling artifacts. To minimize latency and stutter caused by bias, each video (or pair) was fully pre-buffered before the task began. Basic environment checks were enforced by the platform (desktop/laptop and commonly used browsers only). Participants were presented with 12 videos (in pairs) viewed sequentially, which included two verification questions with obvious answers used for participant filtration. Responses from participants who failed either verification comparison were excluded from subsequent analysis. The videos were looped while all user controls were disabled.

\subsection{Pairwise rankings}
To reduce the amount of compared pairs and simplify the comparison process, the dataset was partitioned into subsets consisting of a single uncompressed reference video and all its associated compressed versions for specific presets. For each pair, participants were instructed to select the video exhibiting superior visual quality or to indicate if both videos were of equivalent quality. A minimum of 10 independent responses was collected for each video pair. In total, approximately 1,797,310 valid responses were collected from more than 15,000 unique participants. The application of the online Elo and static Bradley-Terry models to the resulting pairwise ranking data yielded subjective quality scores. Each of these models estimates the underlying preferences based on its own assumptions and optimization framework and may differ slightly in final scores, offering complementary perspectives on the ranking data.
To calculate Elo quality ratings, we followed the methodology provided in \cite{chiang2024chatbot, elo_calc}, which involves comparing multiple videos within the same reference encoding preset and running pairwise battles between them. 
% Every video starts with an initial score, and when video A’s quality is preferred over video B’s quality, video A ``wins'' and takes some points from video B. If a highly-rated video wins against a lower-rated video, the rating change should be smaller because the outcome is expected, and vice versa. The lower-rated player will also gain a few points from the higher-rated player in the event of a draw. This means that this rating system is self-correcting. After numerous iterations, videos with better (for human) perceptual quality will have higher scores than their initial rating. 
To stabilize our online Elo ratings, which recent outliers can skew, we ran the update algorithm on 1,000 bootstrap samples until the confidence intervals ceased to change and then used the median rating across all iterations.
We also applied the Bradley-Terry probabilistic model to convert the pairwise comparison data into ranked subjective quality scores. The resulting subjective scores were normalized for ease of interpretation. 

\subsection{MOS rankings}
\begin{figure*}[tb]
  \includegraphics[width=\textwidth]{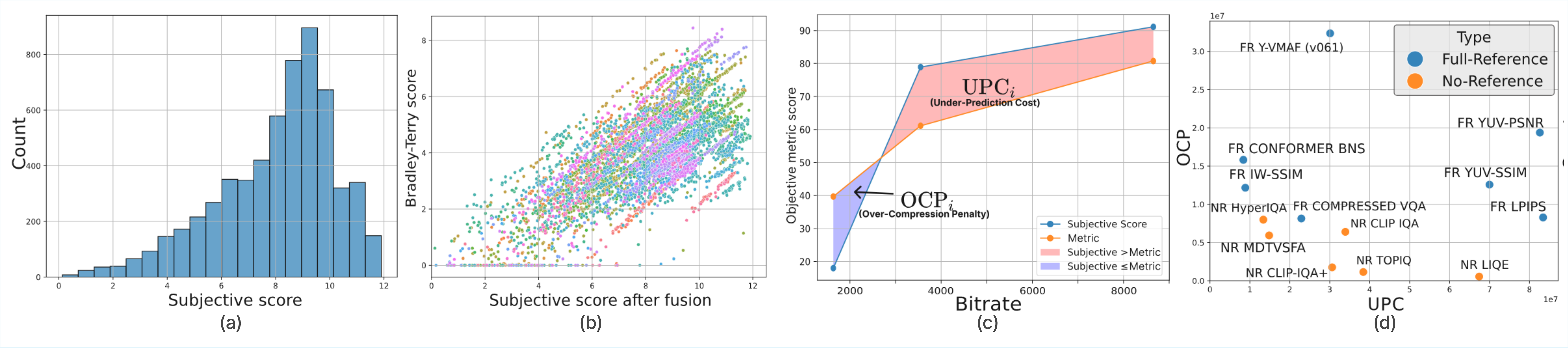}
  \caption{(a) Distribution of MOS in our subjective experiments; (b) Bradley-Terry scores Vs. Fused subjective score; (a) Visualization for $UPC$ and $OCP$ metrics; (d) RDAE results on LEHA-CVQAD, less is better.}
  \Description{LEHA-CVQAD results.}
  \label{fig:example}
\end{figure*}

Subjective scores obtained with pairwise comparisons correlate with MOS, which allows for the simplification of MOSs collection. We used subjective rankings from pairwise comparisons to select a subset of videos, which were then further used for MOS collection. For each group of different reference videos, videos with the highest, median, and lowest quality, as ranked by subjective scores, were selected. A reference video was also added to this pool to enable further DMOS calculations.
To obtain MOSs for these videos, we employed the Absolute Category Rating (ACR) method, a single-stimulus subjective test approach. 
% The continuous scale does not noticeably improve precision \cite{discret1, discret2}; therefore, we used the discrete one for user-friendliness. 
Participants were provided with a sequence to rate its quality on a 21-point discrete scale, where 0 represents the lowest quality, and 20 is the highest. Example of distortions were demonstrated to form participants' rating standards. 
Subjects were instructed to select the point on the scale that best reflected their quality perception with the mouse and keys.
We also included reference videos in the video pool when calculating $DMOS$, which represents the difference between the MOSs of the reference and distorted videos. Before the assessment, participants underwent a training stage to get familiar with the rating interface and video samples. 
We checked if subjects were giving similar quality scores to all videos or was dragging the slider only slightly. As a result, we collected responses from 1496 participants. The distribution of collected MOS values is shown in Fig.~\ref{fig:example}~(a).

To assess the reliability of the collected scores, we performed both inter-subject and intra-subject consistency analyses.
For inter-subject consistency, we randomly split the subjective scores for each video into two equal, non-overlapping groups. MOS was calculated separately for each group, and the Spearman Rank Order Correlation Coefficient (SRCC) was computed between the two sets of MOS values. Repeating this process 100 times with different random splits yielded a median SRCC of 0.992, indicating a high level of internal consistency among subjects.
To evaluate intra-subject consistency, we assessed the reliability of individual subjects’ responses following the procedure described in \cite{subjtest}. The SRCC was calculated between each subject’s opinion scores and the overall MOS. Across all subjects, the median SRCC was 0.973, demonstrating a substantial degree of consistency in individual evaluations.
\subsection{Fusing pairwise and rating scores}

% make sure there is introduction for both rating and pairwise experiment at high level and why we want to fuse them

To gather information from both pairwise and ACR subjective experiments and fuse them, we adapt the method proposed in \cite{mos-bt-fusion}. The original method employed the Thurstone model~\cite {thurston} for pairwise comparisons. The Bradley-Terry~\cite{bradleyterry} model has several advantages, including computational simplicity and speed, robustness to extreme scores% (due to the logistic tail being heavier than the Gaussian tail, which reduces the influence of outliers)
, and improved interpretability. Thus, we adapt the method from \cite{mos-bt-fusion} to the Bradley-Terry model as described below. 
 
In rating subjective experiments, the latent (true) quality $q_i$ can be expressed using the following model of observer rating behavior~\cite{janowski2015accuracy}. For observer $k$ and video $i$, rating $\pi_{ik}$ is $\pi_{ik} = q_i + \delta_k + \xi_{ik}$, where $q_i$ is the ground-truth quality score, $\delta_k$ is the bias of subject $k$ and $\xi_{ik}$ is the subject inaccuracy. In line with prior work\cite{janowski2015accuracy} we assume that all components in the model are independent random variables of normal distribution and $\xi_{ik}$ has zero mean.
% -- $\pi_{ik}$ is also normally distributed. 
Rating scores are usually represented via matrix $M\in\mathbb{R}^{N \times J}$, where $N$ is the number of videos, $J$ is the number of subjects.

The results of pairwise experiments are presented in the form of matrix $C\in\mathbb{N}^{n \times n}$, where $n$ is the number of videos and $c_{ij}$ represents the number of times video $V_i$ was chosen over video $V_j$.
For each video $V_i$ of a true quality $q_i$, a perceived quality $\omega_i \sim Gumbel(q_i, \beta)$  and we can formulate CDF of observer model of Bradley-Terry as:
\begin{equation}
    F_{BT}(\Delta q, \beta) = Logistic(\Delta q_{ij}, \beta) = \frac{1}{1+exp(-\Delta q_{ij} / \beta)}
\end{equation}
Using the Bradley-Terry observer model, the probability of observing matrix $C$ is explained by the Binomial distribution:
\begin{equation}
    P(C\mid q, \beta) =
    \prod_{i<j}
    \binom{n_{ij}}{c_{ij}}\,
    F_{BT}(q_i-q_j,\beta)^{c_{ij}}\,
    \bigl[1-F_{BT}(q_i-q_j,\beta)\bigr]^{n_{ij}-c_{ij}},
\end{equation} where $n_{ij} = c_{ij} + c_{ji}$ -- total number of comparisons for pair $\{i,j\}$.
To keep the same interpretability as in the original paper, we chose the scale $\beta = 1/ln3 \approx 0.9102$. With that choice $F(1,\beta)=0.75$, i.e. a 1-unit gap still means 75 \% of observers prefer the better video. 
Following the assumption of the linear relationship between random variables $\omega_i$ from the pairwise comparison experiment and the random variables $\pi_i$ from the rating experiment, we can state: $\omega_i = a \cdot \pi_i + b \sim Gumbel(q_i, \beta)$,
% \begin{equation}
%     \omega_i = a \cdot \pi_i + b \sim Gumbel(q_i, \beta),
%     \label{eq:omega-with-pi}
% \end{equation} 
$a$ and $b$ are parameters to calculate. This assumption proved sufficient for large-scale datasets \cite{mos-bt-fusion}.

% To account for the potential difference in variability rating scores compared to pairwise choices, we introduce a dispersion factor $c$ that scales the base Gumbel noise parameter $\beta$ to $c\cdot\beta$ in the rating likelihood. By fitting $c$ alongside the affine mapping parameters $(a,b)$, the model can capture this extra observer imprecision in ratings without altering the Bradley–Terry comparison link. 
We can derive the probability density function of the realized value of $\pi_i$ for observer $k$, which we denote by $m_{ik}$:
\begin{equation}
    f(m_{ik}|q_i,a,b,c) = \frac{1}{a c \beta} e^{-z_i-e^{-z_i}}; \quad z_i=\frac{a\,m_{ik}+b-q_i}{c\,\beta}
\end{equation} 
% $\beta$ from Eq.\eqref{eq:omega} becomes $c*\beta$ following the methodology from \cite{mos-bt-fusion}. 
Assuming independence between viewers, the likelihood of observing the whole set of rating scores $M$ is:
\begin{equation}
    P(M|q,a,b,c,\beta) = \prod_{i=1}^{N} \prod_{k=1}^{J} f(m_{ik}|q_i,a,b,c)
\end{equation}
We formulate optimization objective to recover true quality score $q$ from both pairwise and MOS subjective experiments as follows:
\begin{equation}
    \arg\max_{q,a,b,c} P(q,a,b,c|C,M,\beta),
\end{equation} where $P(q,a,b,c|C,M,\beta) \propto P(C|q,\beta) \cdot P(M|q,\beta,a,b,c) \cdot P(q)$; $P(q)$ is a Gaussian prior to regularize the solution:
\begin{equation}
    P(q) = \prod_{i=1}^{N}\frac{1}{\sqrt{2\pi N \sigma^2}} e^{(-\frac{(\mu_q-q_i)^2}{N\sigma^2})},
\end{equation} where $\mu_q$ is mean of quality score $q$.

As a result of fusing procedure, MOS labeled only on a part of the dataset, showed a high correlation with ranking scores across the entire dataset (Fig.~\ref{fig:example}~(b)).

\begin{table*}
  \caption{Results for IQA and VQA metrics on LEHA-CVQAD and its different compression subsets by the proposed bitrate-oriented RDAE metric and by Spearman correlation between quality metric score and subjective score driven by Bradley-Terry and Elo models. For IQA metrics we reported mean value of video per-frame scores.}
  \label{tab:results-1}
  \scalebox{0.91}{%
  \begin{tabular}{clc|ccccccc|cc|cc}
    \toprule
    & & & \multicolumn{7}{c}{RDAE (Gb), MOS $^\downarrow$} & \multicolumn{2}{c}{SRCC, BT $^\uparrow$} & \multicolumn{2}{c}{SRCC, Elo $^\uparrow$} \\
    
    & Model & Type & \makecell{AVC\\ (3 codecs)} & \makecell{HEVC\\ (25 codecs)} & \makecell{VVC\\ (18 codecs)} & \makecell{VP9\\ (2 codecs)} & \makecell{AV1\\ (14 codecs)} & \makecell{Open\\part} & \makecell{Full\\set}& \makecell{Open\\part} & \makecell{Full\\set}& \makecell{Open\\part} & \makecell{Full\\set} \\
    \midrule
    \multirow{3}{*}{\rotatebox[origin=c]{90}{IQA}} &
    PSNR                        & FR & 2.4 & 33.5 & 18.2 & 1.3 & 20.1 
    & 76.3 & 103 & 0.951 & 0.754 & 0.952 & 0.758 \\
    & SSIM    & FR & 2.1 & 30.0 & 21.9 & 1.0 & 27.0 
    & 55.9 & 85.2 & 0.953 & 0.802 & 0.954 & 0.807 \\
    & Linearity\cite{li2020norm}  & NR & 1.5 & 12.9 & 7.4  & 0.5 & 5.2  
    & 24.6 & 38.9 & 0.899 & 0.826 & 0.901 & 0.829 \\
    \multirow{5}{*}{\rotatebox[origin=c]{90}{VQA}} &
    MDTVSFA\cite{li2021unified}     & NR & 1.6 & 10.0 & 8.4  & 0.7 & 3.7  
    & 16.0 & 20.9 & 0.932 & 0.879 & 0.931 & 0.884 \\
    & DOVER\cite{dover}      & NR & 1.7 & 12.2 & 9.3 & 0.8 & 5.4 
    & 17.3 & 24.4 & 0.943 & 0.847 & 0.933 & 0.838 \\
    & Q-Align\cite{qalign}   & NR & 1.9 & 13.0 & 9.5 & 0.8 & 5.7 
    & 19.5 & 26.1 & 0.940 & 0.849 & 0.925 & 0.842 \\
    & Faster-VQA\cite{wu2022fast} & FR & 1.2 & 9.8 & 8.3 & 0.6 & 3.6 
    & 16.4 & 19.6 & 0.939 & 0.841 & 0.926 & 0.837 \\
    & VMAF\cite{li2018vmaf}      & FR & 1.6 & 21.5 & 13.5 & 0.7 & 15.2 
    & 51.8 & 64.5 & 0.952 & 0.853 & 0.954 & 0.856 \\
  \bottomrule
\end{tabular}
}
\end{table*}

\begin{table*}
  \caption{Correlations of quality assessment metrics with MOS on LEHA-CVQAD (open part) and various other datasets.}
  \label{tab:results-2}
  \scalebox{0.88}{%
  \begin{tabular}{lcccccccccccccc}
    \toprule
    
    Model & \multicolumn{2}{c}{\textbf{LEHA-CVQAD}} & \multicolumn{2}{c}{CVQAD\cite{antsiferova2022video}} & \multicolumn{2}{c}{LSVQ\cite{lsvq}} & \multicolumn{2}{c}{YT-UGC\cite{ytugc}} & \multicolumn{2}{c}{KoNViD-1k\cite{konvid}} & \multicolumn{2}{c}{LIVE-VQC\cite{livevqc}}  & \multicolumn{2}{c}{KUGVD\cite{kugvd}} \\
     & SRCC$^\uparrow$ & PLCC$^\uparrow$ & SRCC$^\uparrow$ & PLCC$^\uparrow$ & SRCC$^\uparrow$ & PLCC$^\uparrow$ & SRCC$^\uparrow$ & PLCC$^\uparrow$ & SRCC$^\uparrow$ & PLCC$^\uparrow$ & SRCC$^\uparrow$ & PLCC$^\uparrow$ & SRCC$^\uparrow$ & PLCC$^\uparrow$ \\
    \midrule
    Linearity\cite{li2020norm}  & 0.439 & 0.432 & 0.876 & 0.862 
    & 0.484 & 0.491 & 0.315 & 0.332 & 0.720 & 0.734 
    & 0.718 & 0.713  & 0.743 & 0.736 \\
    MDTVSFA\cite{li2021unified}   & 0.639 & 0.635 & 0.929 & 0.943 
    & 0.545 & 0.551 & 0.445 & 0.454 & 0.829 & 0.822 
    & 0.880 & 0.869  & 0.726 & 0.734 \\
    DOVER\cite{dover}    & 0.761 & 0.757 & 0.921 & 0.923 
    & 0.884 & 0.879 & 0.884 & 0.876 & 0.883 & 0.878 
    & 0.836 & 0.845 &  0.824 & 0.826 \\
    Q-Align\cite{qalign}   & 0.768 & 0.761 & 0.876 & 0.891 
    & 0.889 & 0.886 & 0.891 & 0.893 & 0.894 & 0.886 
    & 0.893 & 0.884  & 0.833 & 0.838 \\
    Faster-VQA\cite{wu2022fast} & 0.746 & 0.739 & 0.845 & 0.859 
    & 0.857 & 0.854 & 0.873 & 0.874 & 0.863 & 0.857 
    & 0.824 & 0.830  & 0.801 & 0.807 \\
    VMAF\cite{li2018vmaf}       & 0.755 & 0.773 & 0.943 & 0.952 
    & ---   & ---   & ---   & ---  & --- & ---  
    & --- & --- & 0.952 & 0.954 \\
  \bottomrule
\end{tabular}
}
\end{table*}

\section{Experiments and results}
\subsection{Dataset diversity}
For videos within LEHA-CVQAD, as well as for other publicly available datasets, we computed and analyzed several spatial and temporal statistical measures, including the following: Spatial Information (SI) \cite{complex_si}, Temporal Information (TI) \cite{complex_ti}, Contrast \cite{konvid}, Colorfulness (CF) \cite{complex_cf}, Luminance \cite{complex_abs}, Score \cite{complex_sharpness}, and IC9600 complexity metric \cite{complex_ic9600}. Fig.~\ref{fig:complexity} illustrates that range and uniformity characteristics of calculated features for LEHA-CVQAD outperform other datasets, including CVQAD, and are even comparable with UGC datasets.

% Spatial Information (SI), indicating
% the amount of local spatial variation in each frame, (ii)
% Temporal Information (TI), which captures change across
% frames, and the (iii) Colorfulness (CF) measure. SI is
% a Sobel magnitude measure, whereas TI uses the average
% squared luminance difference between successive frames:

\subsection{Benchmark methodology}
We launched a benchmark of 100 various IQA/VQA metrics on the proposed LEHA-CVQAD dataset. A subset of metrics was used to compare the correlations with MOS on LEHA-CVQAD with correlations on other datasets.

\textbf{Standard evaluation metrics}.
We adopted a standardized protocol to compare predicted scores with ground-truth subjective ratings, estimating each method's monotonicity using Spearman's rank correlation coefficient (SRCC) and its prediction accuracy using Pearson's linear correlation coefficient (PLCC). In the case of MOS, we calculated correlation coefficients across the entire dataset simultaneously. For ranking-based subjective scores, we computed correlations for each group, corresponding to the reference video and encoder preset, separately. 
% Moreover, SRCC and PLCC were computed only for groups with at least 12 and 6 samples, correspondingly. 
To derive a single overall coefficient per method, we applied the Fisher Z‑transformation \cite{fisher_transformation} to each group result, calculated a sample‑size‑weighted mean of those transformed values along with confidence intervals, and then used the inverse transform to return to the correlation scale. We used DMOS instead of MOS for full-reference metrics.

% A standardized evaluation protocol was adopted to assess the performance of each quality assessment method from two complementary perspectives: monotonicity and prediction accuracy. Monotonic relationships between the predicted quality scores and ground-truth subjective ratings were quantified using the Spearman Rank-Order Correlation Coefficient (SRCC) and the Kendall Rank-Order Correlation Coefficient (KROCC). In parallel, the Pearson Linear Correlation Coefficient (PLCC) was employed to evaluate the accuracy of the predictions by measuring the strength of the linear association between predicted and subjective scores.

% For each combination of reference video and encoder preset, SRCC, KROCC, and PLCC were computed between the predictions of each quality assessment method and the corresponding subjective scores. To ensure the robustness and statistical reliability of the correlation estimates, only groups with at least 15 samples were considered for SRCC, and those with at least six samples for KROCC and PLCC.

% Correlation coefficients were subsequently aggregated by first applying the Fisher Z-transformation (inverse hyperbolic tangent) to each group’s result, weighting the transformed scores proportional to group size. The resulting weighted average was then mapped back to the correlation scale via the inverse Fisher Z-transform, yielding a single representative correlation coefficient for the entire dataset \cite{fisher_transormation}.

\textbf{A new bitrate-saving-oriented metric}.
% When a practical encoder parameters selection is driven by a quality metric $M$, it chooses the quantization step size (or $\lambda$) that achieves a target value $q$ on $M$ for every bitrate. If $M$ underestimates the true perceptual quality, the encoder compresses more lightly to meet $q$; if it overestimates, it delivers a smaller file at the cost of visible artifacts. RDAE measures the accumulated mismatch between the objective and subjective rate-distortion (RD) curves, thereby answering the engineering question: ``How many bits or just noticeable differences could I lose by trusting this metric?''.
Video encoder settings are often tuned to achieve a target real perceptual quality $q$ using VQA metric $M$. For example, if a content provider wants a video quality of ``8 out of 10'', they may use SSIM and set its threshold for compressed videos to 0.8. Underestimating perceptual quality by an objective metric leads to over-compression and more disturbing artifacts. When $M$ overestimates the actual quality of a video, it results in an unnecessarily large compressed video file size and streaming costs. To calculate the amount of metric error in this setting, we propose a \textit{Rate–Distortion Alignment Error (RDAE)} that measures how well an objective VQA metric can serve as a proxy for perceptual quality during encoder parameter tuning, answering: ``How many bits could be lost by trusting $M$?''.

RDAE consists of two components: Under-prediction Cost (UPC) and Over-compression Penalty (OCP), illustrated in Fig.~\ref{fig:example}~(c).
The calculation of UPC, OCP, and RDAE is relatively straightforward. First, all objective scores and MOS values are mapped onto a shared perceptual scale using Neural Optimal Transport~\cite{not}. 
    % Common perceptual scale. All objective scores and MOS values are first mapped onto a shared perceptual scale via a Neural Optimal Transport \textbf{not}.
Second, data is partitioned into disjoint groups (same source, codec, preset) with at least three bitrate points.
    % Formation of RD groups. The dataset is partitioned into disjoint groups $g_i$ that contain the same source sequence, codec, and preset, while spanning at least three bitrate points. Thus, we can plot Rate–Distortion (RD) curve with bitrate on X-axis and Subjective score/metric value on Y-axis. Example can be seen on \ref{fig:example}.
For each group $i$, let $s_i(b)$ be the subjective score and $\hat s_i(b)$ the metric value at bitrate $b$, then:
    % Calculate $UPC_i$ and $OCP_i$ for each group. We connect adjacent points on the aforementioned plot and calculate two areas between the subjective ($s(b)$) and objective ($\hat{s}(b)$) RD curves. 
    \begin{align}
    \mathrm{UPC_i}/\mathrm{OCP_i} &= 
        \int_{\{s(b) \geq/< \hat{s}(b)\}}[s(b) - \hat{s}(b)]db.
    % \\
    % \mathrm{OCP_i} &= 
    %     -\int_{\{s(b) < \hat{s}(b)\}}[s(b)-\hat{s}(b)] db,
    \end{align}
Finally, the UPC and OCP scores for each group are averaged across the entire dataset, and RDAE is calculated as a sum of these scores.
    % \begin{align}
    %     \mathrm{UPC} = \sum_{i}\mathrm{UPC_i}; \mathrm{OCP} = \sum_{i}\mathrm{OCP_i}; \mathrm{RDAE = UPC + OCP}
    % \end{align}
%Commonly used Pearson, Spearman, and Kendall correlations rank metrics are fundamental indicators of VQA metric's performance, but they (i) ignore the direction of error (bits wasted vs. quality lost), (ii) treat each point on RD curve independently even though encoders traverse the curve continuously, (iii) are insensitive to error magnitude, and (iv) provide no actionable unit. RDAE resolves all four issues by producing interpretable areas expressed in ``perceptual-points $\times$ log-bitrate`` units. 
The proposed metric and its components can be viewed as indicators of how well the IQA/VQA metric optimizes the codec to save bitrate (UPC) or how effectively it helps preserve perceived quality (OCP).
%The proposed RDAE family, therefore, complements classical accuracy statistics with a direct, interpretable measure of operational risk, better reflecting the needs of bitrate-constrained video delivery systems.

\subsection{Benchmark results}
The results in Table~\ref{tab:results-1} show that many widely used metrics, including PSNR, SSIM, and VMAF, exhibit high RDAE values.
% , particularly for videos compressed with codecs of new standards, such as AV1 and VVC. 
This indicates potential inefficiencies in bitrate allocation when relying solely on these metrics during video transcoding optimization. MDTVSFA\cite{li2021unified} show promising results for this task, as illustrated in Fig.~\ref{fig:example}~(d).

Table~\ref{tab:results-2} shows correlation results on different datasets. Top-performing metrics achieve lower SRCC and PLCC values on LEHA-CVQAD MOS scores than on other datasets, indicating that overall dataset complexity provided by various compression, hidden proprietary codecs, and UGC content remains challenging for current models.

\section{Conclusion}

In this paper, we introduced LEHA-CVQAD, a new large-scale dataset with multi-modal quality labels for compressed video quality assessment. LEHA-CVQAD contains 59 videos of different genres compressed using 186 distinct video codecs across three target bitrates, resulting in 6,240 distorted videos. We performed a thorough complexity analysis and demonstrated that LEHA-CVQAD surpasses existing datasets, such as CVQAD and other publicly available collections, in terms of content diversity, codec variety, and distortion realism. In addition to pairwise subjective comparisons yielding BT and ELO scores, we collected MOS to facilitate the usage of LEHA-CVQAD for training VQA metrics. 
Finally, we evaluated 100 IQA/VQA metrics and demonstrated that the vast majority failed to capture the range of compression artifacts comprehensively. We believe that LEHA-CVQAD will serve as a robust, fair, and challenging evaluation platform, inspiring the development of new video quality assessment methods that can handle contemporary compression artifacts. Future work in this field requires expanding the dataset to incorporate additional resolutions, such as 4K and 8K, higher bit-depth, and content types.
    
% \begin{table}
%   \caption{Frequency of Special Characters}
%   \label{tab:freq}
%   \begin{tabular}{ccl}
%     \toprule
%     Non-English or Math&Frequency&Comments\\
%     \midrule
%     \O & 1 in 1,000& For Swedish names\\
%     $\pi$ & 1 in 5& Common in math\\
%     \$ & 4 in 5 & Used in business\\
%     $\Psi^2_1$ & 1 in 40,000& Unexplained usage\\
%   \bottomrule
% \end{tabular}
% \end{table}

% \begin{figure}[ht]
%   \centering
%   \includegraphics[width=\linewidth]{imgs/sample-franklin.png}
%   \caption{1907 Franklin Model D roadster. Photograph by Harris \&
%     Ewing, Inc. [Public domain], via Wikimedia
%     Commons. (\url{https://goo.gl/VLCRBB}).}
%   \Description{A woman and a girl in white dresses sit in an open car.}
% \end{figure}

%\cite{Cohen07}

%%
%% The acknowledgments section is defined using the "acks" environment
%% (and NOT an unnumbered section). This ensures the proper
%% identification of the section in the article metadata, and the
%% consistent spelling of the heading.
\begin{acks}
The work was supported by a Research Center for Trusted Artificial Intelligence the Ivannikov Institute for System Programming of the Russian Academy of Sciences.
The research was carried out using the MSU-270 supercomputer of Lomonosov Moscow State University. We also would like to express our gratitude to MSU Video Codecs Comparisons team for discussing the results of this research.
\end{acks}

%%
%% The next two lines define the bibliography style to be used, and
%% the bibliography file.
\bibliographystyle{ACM-Reference-Format}
\bibliography{sample-base}

% %%
% %% If your work has an appendix, this is the place to put it.
% \appendix

% \section{Results with confidence intervals}

% \section{Statistical tests}

\end{document}